\documentclass[journal]{IEEEtran}

\ifCLASSINFOpdf
  \usepackage[pdftex]{graphicx}
\else
  \usepackage[dvips]{graphicx}
\fi

\usepackage{makeidx}  
\usepackage{amsmath}
\usepackage{booktabs}
\usepackage{tabularx}
\usepackage{lipsum}
\usepackage{multirow}

\begin{document}
%
\title{Morphometry-Based Longitudinal Neurodegeneration Simulation with MR Imaging}
\author{Siqi~Liu,
        Sidong~Liu,
        Sonia~Pujol,
        Ron~Kikinis,
        Dagan~Feng,
        Michael~Fulham,
        and~Weidong~Cai
\thanks{SQ. Liu, SD. Liu, D. Feng, M. Fulham, W. Cai are with School of Information Technologies, The University of Sydney, Sydney, NSW, Australia}
\thanks{M. Fulham is also with PET and Neuclear Medicine Department, Royal Prince Alfred Hospital, Sydney, NSW, Australia.}
\thanks{S.Pujol and R. Kikinis are with Surgical Planning Laboratory, Harvard Medical School, Boston, MA, United States}
}


\maketitle

\begin{abstract}
We present a longitudinal MR simulation framework which simulates the future neurodegenerative progression by outputting the predicted follow-up MR image and the voxel-based morphometry (VBM) map. This framework expects the patients to have at least 2 historical MR images available. The longitudinal and cross-sectional VBM maps are extracted to measure the affinity between the target subject and the template subjects collected for simulation. Then the follow-up simulation is performed by resampling the latest available target MR image with a weighted sum of non-linear transformations derived from the best-matched templates. The leave-one-out strategy was used to compare different simulation methods. Compared to the state-of-the-art voxel-based method, our proposed morphometry-based simulation achieves better accuracy in most cases.
\end{abstract}

\begin{IEEEkeywords}
Neurodegenerative Disease, Registration, MRI.
\end{IEEEkeywords}

%
\IEEEpeerreviewmaketitle

\section{Introduction}
Neurodegenerative diseases are a category of progressive loss of structure or function of neurons, including Alzheimer's Disease (AD), Parkinson's disease (PD), Huntington's disease (HD) and Spinal muscular atrophy (SMA), etc. Neuroimaging has advanced the analysis of neurodegenerative processes profoundly in the past two decades with large-scale group studies \cite{sdreview1, shen2014genetic, sdreview2}. The large-scale neuroimaging computing methods of neurodegenerative diseases can be generally categorised into cross-sectional and longitudinal. Though the majority of the recent neuroimaging studies focused on the cross-sectional group comparison with regional measurements \cite{sdmbk, sdpgf, sqisbi14, sqtbme},
the longitudinal analysis of brain tissue changes is effective in describing an evolving neurodegenerative process \cite{sqisbi15, scahill2003longitudinal,ho2003progressive}. A number of investigations have attempted to analyse the longitudinal changes from neuroimaging biomarkers at serial time-points and to make predictions of the underlying process \cite{Davatzikos20011361,HBM:HBM22511,Sabuncu20149}. However, it is difficult to evaluate any such predictions in practice due to the lack of ground truth.  We suggest that simulation of  follow-up brain images can be useful for validating the predictions obtained from neuroimaging biomarkers. 

The longitudinal simulation of the follow-up structural magnetic resonance (sMR) images can be performed by resampling the baseline sMR image with a weighted average of the longitudinal non-linear transformations from a template population \cite{modat2014simulating}. When the template population is sufficiently large to include a majority of neurodegenerative changes, the confidence in simulated results would heavily depend on the metrics used to measure the affinity between the target patient and the template population.

Previous frameworks of longitudinal sMR simulation generally depended on two assumptions about the progress of neurodegeneration: (1) a common rate of atrophy is shared across all subjects with the same diagnostic label; (2) brains with similar morphology evolve in a similar way \cite{Davatzikos20011361,sharma2013estimation,modat2014simulating}.
Alzheimer's disease (AD), which is the most common neurodegenerative disorder, progresses in a special pattern which has multiple stages. The neurofibrillary tangles (NFT), which are thought to contribute to local atrophy, spread from memory related areas towards areas in the medial temporal lobe, the parietal cortex and the prefrontal cortex \cite{braak2011stages}. Thus, patients in different stages of progression stages might not have similar atrophy rates across all brain regions. There might also be difficulty in comparing the local structural morphology accurately between different patients only based on the sMR intensities due to the inter-subject variance in the original structural appearances. Besides the sMR intensities, the historical progression of the same patient (longitudinal) as well as the difference between the current state of the patient and a normal brain template may generate the simulation from an alternative perspective.

\begin{figure*}[!htb]
\centering
\includegraphics[width=1\textwidth]{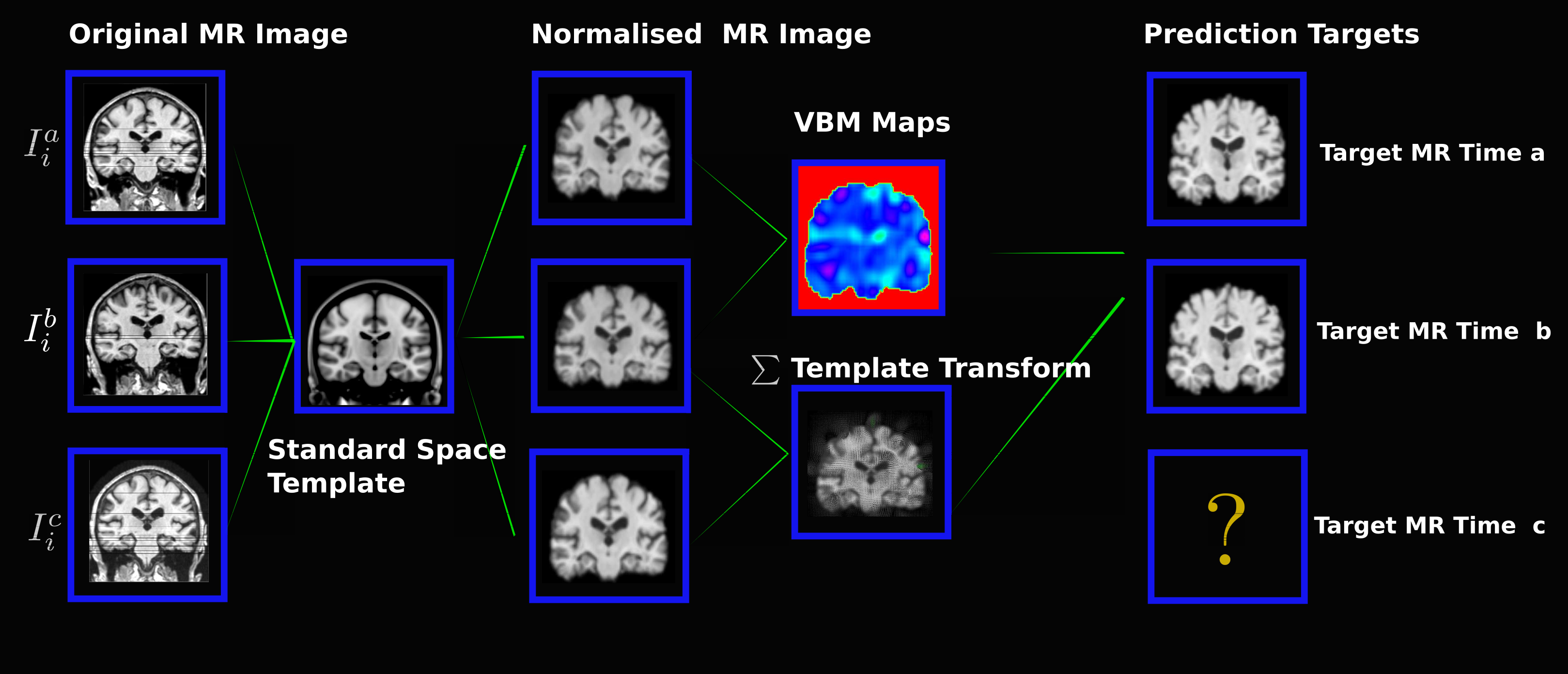}
\caption{An illustration of the proposed framework of the longitudinal MR simulation. }
\label{fig:framework}
\end{figure*}

In this study, we present a proof-of-concept framework to simulate  neurodegeneration in follow-up with sMR data. An overview of the framework is illustrated in Fig.~\ref{fig:framework}. We hypothesise that brains with similar detected cross-sectional and longitudinal atrophic deformations would have similar follow-up evolution. The cross-sectional changes are derived by a symmetric non-linear registration from a standard template to the subjects; the longitudinal registration between two serial sMR images of the same patient is used to extract the intra-subject serial changes. We apply the voxel-based morphometry (VBM) to measure both types of structural brain deformations. Each subject recruited in the template population is expected to have at least 3 serial sMR visits. Both the longitudinal VBM map and the cross-sectional VBM map are collected for each template subject. The simulation of the future sMR volume of a target patient is performed by resampling the latest available sMR volume with an average weighted sum of the longitudinal transformations in the following period in the template population. Only a minority of the best-matched templates in both VBM maps are selected to contribute to the simulation results. The cut-off threshold depends on the size of the recruited template population. 

%
%
%
%

 

\section{Methods}
\subsection{Preprocessing}
Each template subject $S_i$ is expected to have at least 3 serial sMR visits taken at time-points $a,b,c$ and $a<b<c \wedge b-a \approx c-b$. Each template MR volume is skull-stripped using the Brain Extraction Tool with a standard space pre-masking applied \cite{HBM:HBM10062}. Then all template MR images are affine registered to the MNI152 template brain space with FLIRT \cite{Jenkinson2002825}. A symmetric diffeometric registration is performed between each pair of the standardised MR images $[I_i^{(a)}, I_i^{(b)}]$ and $[I_i^{(b)}, I_i^{(c)}]$ to obtain the longitudinal brain tissue deformations $T_i^{(ab)}$ and $T_i^{(bc)}$. The symmetric registration firstly registers the two adjacent MR volumes to a mid-way space to ensure the inverse consistency of the forward and backward transformations which was known to be important for atrophy calculation \cite{leung2012consistent,christensen2001consistent}. The MNI152 template $M$ is symmetrically registered to each $I_i^{(b)}$ image to obtain the cross-sectional deformation $T_i^{(Mb)}$.

\subsection{Template Weighting with Voxel-Based Morphometry}
The voxel-based morphometry (VBM) maps \cite{ashburner2000voxel,chung2001unified} were calculated on $T_i^{(ab)}$ and $T_i^{(tb)}$ respectively as the Jacobian determinant of the spatial transforms. $U$ is the displacement of $T$. The displacement tensor of $U$ over time $t$ is represented as 
\begin{equation}
\frac{\partial U}{\partial x}(x, t)=
  \begin{pmatrix}
    \frac{\partial U_1}{\partial x_1} & \frac{\partial U_1}{\partial x_2} & \frac{\partial U_1}{\partial x_3} \\
    \frac{\partial U_2}{\partial x_1} & \frac{\partial U_2}{\partial x_2} & \frac{\partial U_2}{\partial x_3} \\
    \frac{\partial U_3}{\partial x_1} & \frac{\partial U_3}{\partial x_2} & \frac{\partial U_3}{\partial x_3}
  \end{pmatrix}
\end{equation}

The determinant $det(\frac{\partial U}{\partial x}(x, t))=det(I+\nabla U)$ is calculated at each voxel and forms the VBMs. To measure the distance between two detected tissue deformation $T_i$ and $T_j$, the squared Euclidean distance of the VBMs is calculated within the area of a dilated brain mask $\wp$ in the standardised space as $D(J_i, J_j) = \sum_{v \in \wp}{(d_v^{(i)} - d_v^{(j)})^2}/|\wp|$, where $d_v^{(i)}$ is the Jacobian determinant of $T_i$ at voxel $v$. The area near the outer boundary of the dilated brain mask is used to capture the deformation on or near the cortical surface.

\begin{table*}[t]
     \begin{center}
     \caption{The visual check of a successful simulation and a failed simulation. The column Year 1 is the MR image at the second time-point; the column Year 2 is the real follow-up MR image of this patient and the column Simulated is the predicted Year 2 MR image. The VBM maps are overlaid on the Year 2 and Simulated images. The failure was probably caused by the spatial mismatch introduced in the automatic affine normalisation.}
     \bgroup
     \begin{tabular}{c c  c  c}
     \toprule
       & Year 1 & Year 2 & Simulated \\ 
     \hline
     \multirow{14}*{Successful} 
     &
     \raisebox{-\totalheight}{\includegraphics[width=0.28\textwidth]{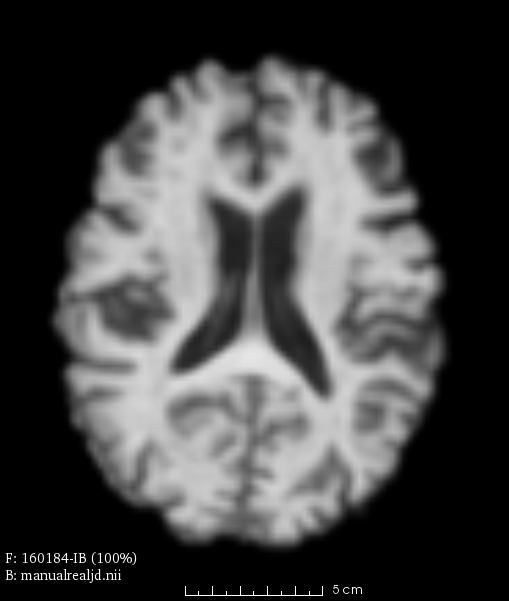}}
      & 
      \raisebox{-\totalheight}{\includegraphics[width=0.28\textwidth]{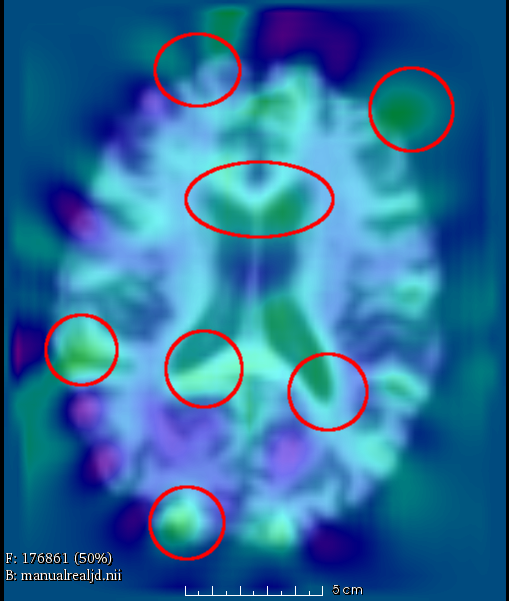}}
      &
      \raisebox{-\totalheight}{\includegraphics[width=0.28\textwidth]{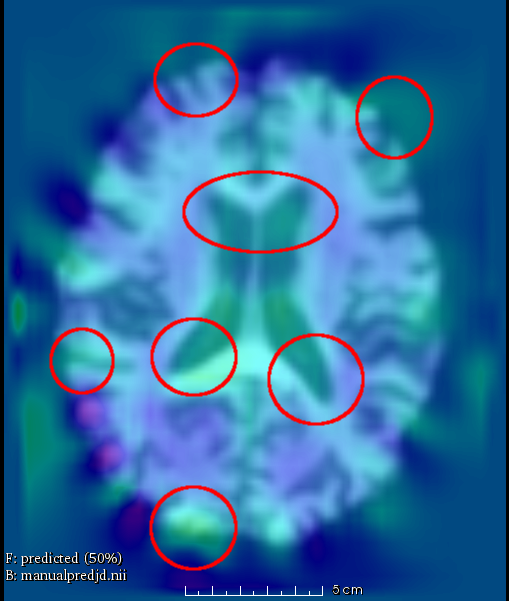}}\\\hline
      \multirow{14}*{Failed} 
     &
     \raisebox{-\totalheight}{\includegraphics[width=0.28\textwidth]{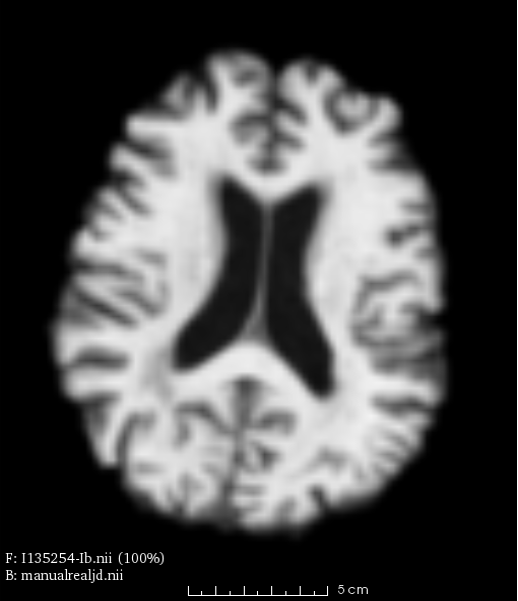}}
      & 
      \raisebox{-\totalheight}{\includegraphics[width=0.28\textwidth]{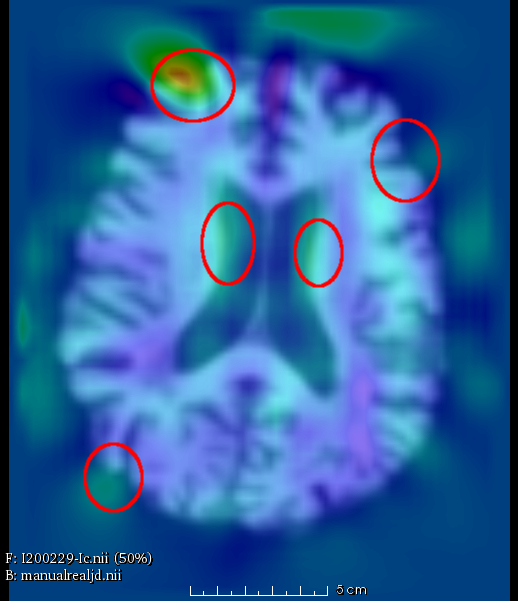}}
     &
     \raisebox{-\totalheight}{\includegraphics[width=0.28\textwidth]{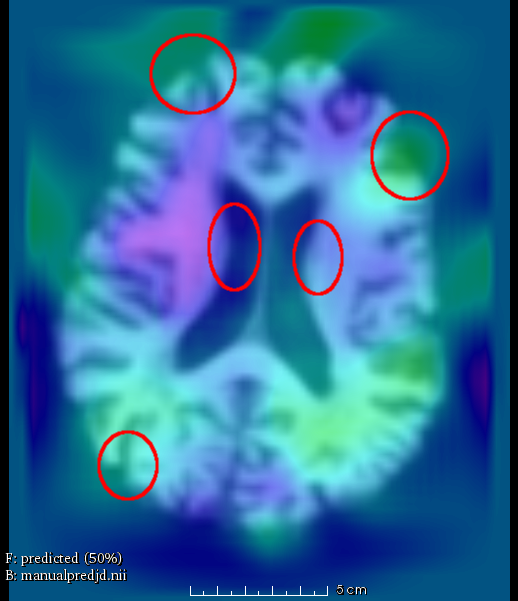}} \\
      \bottomrule
      \end{tabular}
     \egroup 
      \\
      
      \label{fig:visualcheck}
      \end{center}
\end{table*}

\subsection{Follow-Up MR Simulation}
To simulate the follow-up MR volume $I_\star^c$ of the target patient $S_\star$, we require the patient to have two MR volumes available $I_\star^a$ and $I_\star^b$ and the interval $(b-a)$ to approximately equal the intervals in the templates. The symmetric longitudinal and cross-sectional deformations are obtained as $T_\star^{ab}$ and $T_\star^{Mb}$. The VBMs of both transformations are correspondingly computed as $J_\star^{ab}$ and $J_\star^{Mb}$ which are used for collecting the masked longitudinal distances $D_i^{long}(\star,i)=D(J_\star^{ab}, J_i^{ab})$ and the cross-sectional distances $D_i^{cross}(\star,i)=D(J_\star^{Mb}, J_i^{Mb})$. Two types of collected distances are respectively normalised within [-1, 1] as $\tilde{D}=(D-\bar{D})/max(D-\bar{D})$
where $\bar{D}$ is the mean value of $D$. Then both distances are summed with relative weights to obtain the combined  $D_i^{\prime}=\alpha \bar{D}_i^{long} + \beta \bar{D}_i^{cross}$, $\alpha+\beta=1$. k nearest neighbours of $S_\star$ are selected to form a new template set K. The majority of the templates are dropped out for one simulation because the distant subjects would introduce bias rather than contributing to the simulation accuracy, even Gaussian distributed weights are used. Based on $D_i^{\prime}$, the follow-up deformation $T_\star^bc$ of subject $S_\star$ is computed as an average weighted sum of the template transformations $T_i^{bc}$ as
\begin{equation}
T_\star^{bc}  = \frac
{\sum_{i\in |K|}{T_i^{a\star} \times e^{-D^\prime_i / g}}}
{\sum_{i\in |K|}{e^{-D^\prime_i / g}}}
\end{equation}
where $T_i^{a\star}$ is the forward transformation from the template image $I_i^a$ to the target image $I_\star^a$; $g$ is the Gaussian kernel density which is set to 0.5 \cite{modat2014simulating}. The follow-up image $I_\star^c$ can be simulated by resampling $I_\star^b$ with $T_\star^bc$. Notably, in this proposed framework, subjects with different diagnostic labels are not required to be computed separately, since there might not be clear boundaries between the atrophic progression patterns from different diagnostic labels. For example, some follow-up transformations of early AD patients may contribute to simulate the future neurodegeneration of a late MCI patient. With the drop-out threshold $|K|$, the distant subjects are expected to be filtered out before the template merging.

\begin{figure*}[!htb]
\centering
\includegraphics[width=0.9\textwidth]{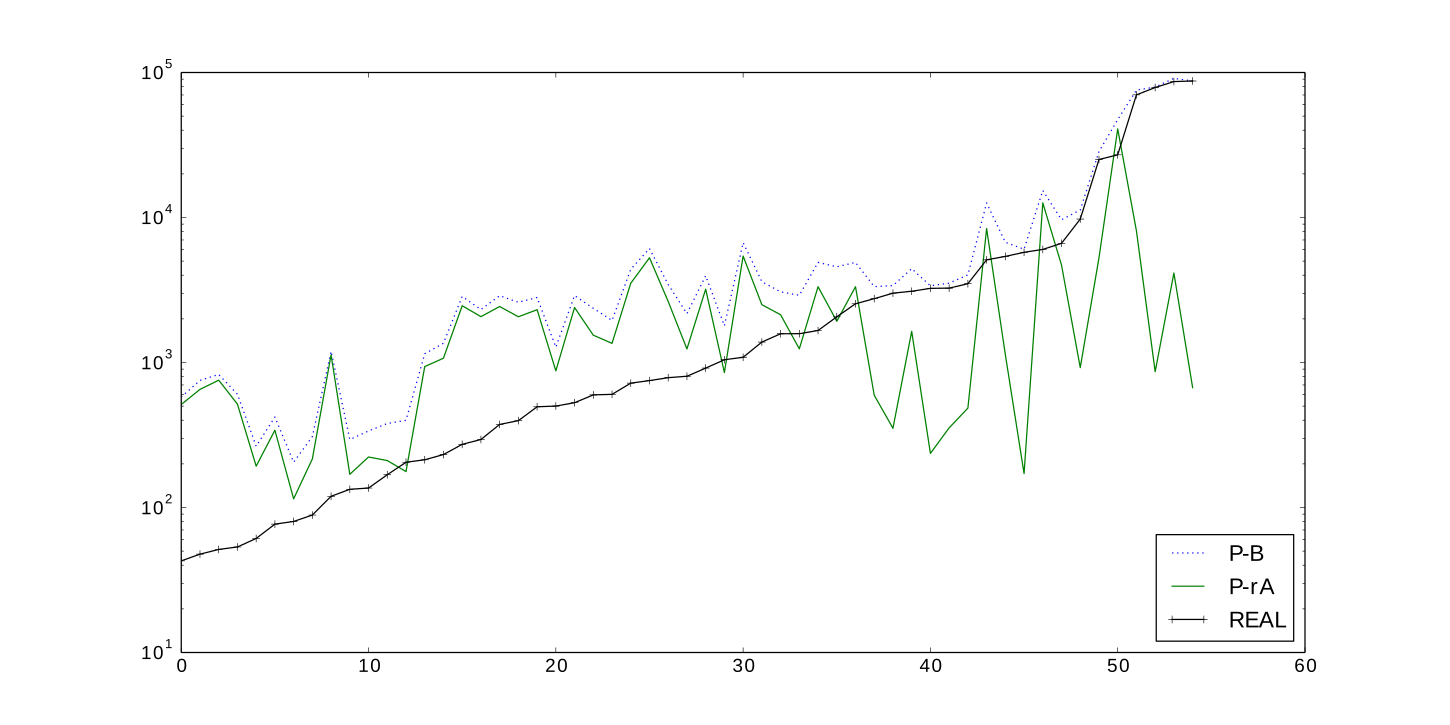}
\caption{The plot of the three types of MR image distances considered for this evaluation: The distances between the predicted images and the real follow-ups (P-B); The distances between the predicted images and the registered 1 year MR images (P-rA); The distances between the registered 1 year MR images and the real follow-up MR images (REAL) }
\label{fig:3errors}
\end{figure*}

\section{Evaluation and Results}

\begin{figure*}[!htb]
\centering
\begin{minipage}[b]{1\textwidth}
  \centering
  \centerline{\includegraphics[width=1\linewidth]{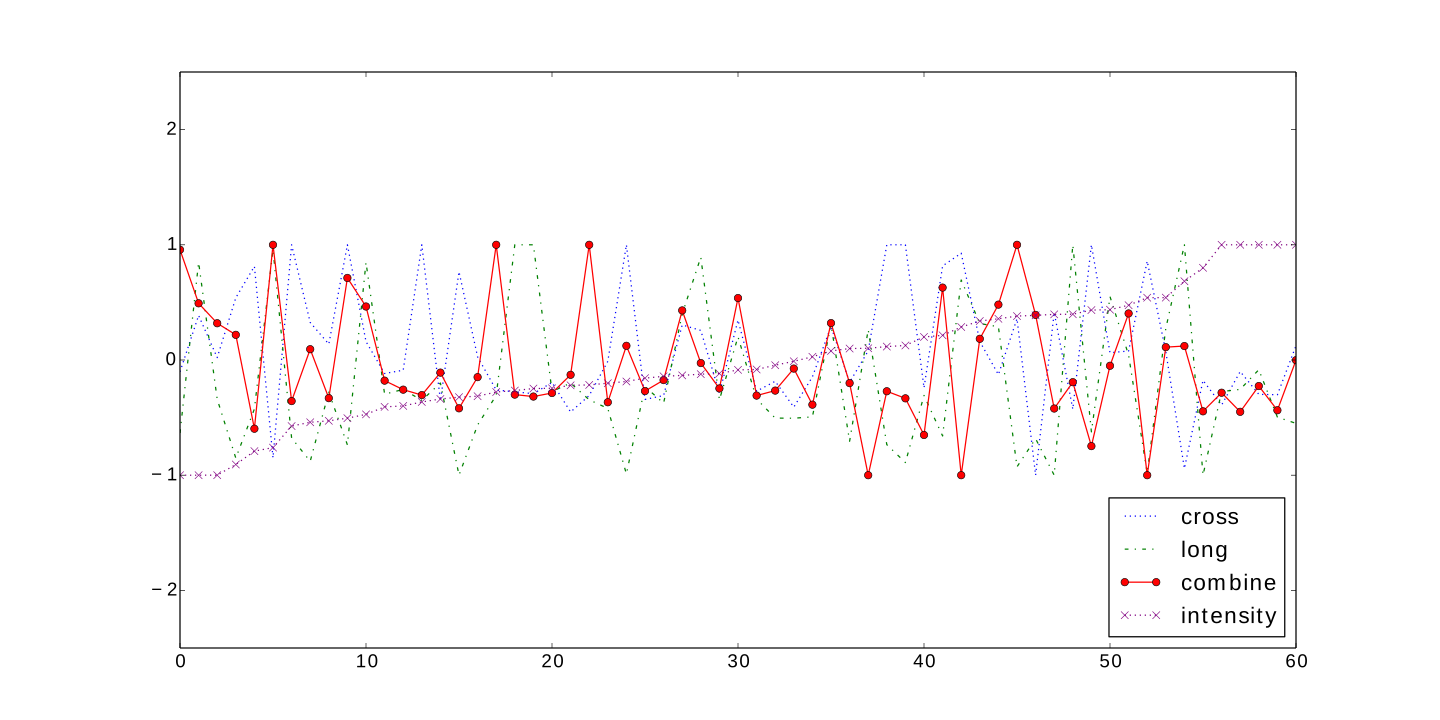}}
  \centerline{(a) Sorted distances between the predicted MR and the real follow-up MR (P-B)}\medskip
\end{minipage}
\hfill
\begin{minipage}[b]{1\textwidth}
  \centering
  \centerline{\includegraphics[width=1\linewidth]{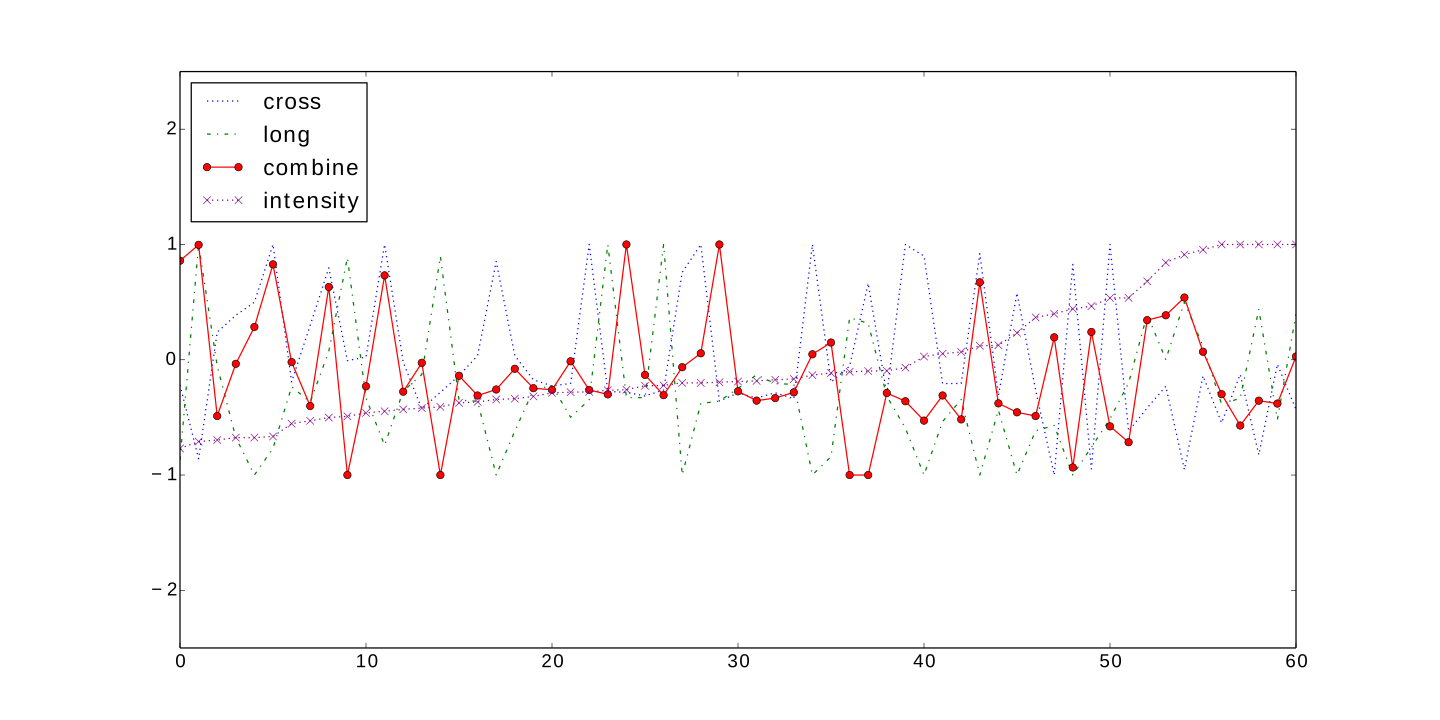}}
  \centerline{(b) Sorted distances between the predicted and the registered Year 1 (P-rA)}\medskip
\end{minipage}
\caption{The plot of different template weighting methods: cross-sectional VBM weighting (cross), longitudinal VBM weighting (long), combined weighting (combine), intensity-based weighting (intensity). All distances were zero-mean rescaled into $[-1,1]$ and the subjects were sorted according to the intensity-based weighting.  Lower values indicate better simulation results.}
\label{fig:compare}
\end{figure*}

We recruited GradWarped MR images with N3 Correction from the publicly available ADNI 1 and ADNI GO dataset (http://adni.loni.usc.edu/) \cite{jack2010update}. The slice thickness of all MR volumes is 1.3mm. We kept 60 subjects who had at least three continuous MR visits with an interval of 12 months available, resulting in 180 MR volumes and 120 MR longitudinal pairs in total to be registered. After transforming each MR volume to the standard MNI152 space, skull stripping was conducted with a fraction threshold of 0.3. We used the MNI152 template with 2mm slice thickness for the image normalisation.

We used the leave-one-out strategy to evaluate the framework. The $I^a$ and $I^b$ of each selected patient were used as the inputs to simulate the unknown follow-up $I^c$. The rest of the patients were used as the simulation templates. All the transformations and images belong to the testing target are excluded from the template construction. The simulated $I^c$ is then compared with the real follow-up image as well as the image derived by registering $I^b$ to the real follow-up image.

The symmetric non-linear registration was implemented with the Advanced Normalization Tools (ANTS) \cite{Avants20112033}. The integration of the entire framework was implemented based on the Nipype framework \cite{gorgolewski2011nipype}.

Examples of a successful simulation on an AD patient and a failed simulation of a normal control patient are presented in Table.~\ref{fig:visualcheck}. In each case, the Year 1 MR image ($I^b$) is shown as a reference; The real follow-up Year 2 MR image ($I^c$)  and the simulated follow-up MR image ($I^c_\star$) are overlaid on the VBM map extracted between them and the Year 1 MR image to visualise the longitudinal deformations. Comparing the simulated MR image and the real follow-up of the successful case, the tendency for the ventricles to increase in size, reflecting more cortical atrophy was successfully simulated. The VBM map also showed an approximately matched tissue deformation along the ventricle as well as the cortical areas. The example that failed showed a mismatch of the atrophy in the MR images and the VBM maps. The increase in size of the lateral ventricles was neglected and the cortical atrophy was overestimated. Such simulation failure can be generally attributed to the affine normalisation errors and the insufficiency of the longitudinal template collection.

To evaluate the strengths of different template weighting strategies, we considered the square distances between the simulated volume and the real follow-up volume (P-B) as well as the distances between the simulated volume and the registered Year 1 volume (P-rA), which is the output of registering the Year 1 image to its real follow-up MR image (Fig.~\ref{fig:3errors}). In Fig.~\ref{fig:3errors}, unlike P-rA, the P-B curve is correlated with the original registration error. It  might indicate that P-rA could be a more unbiased evaluation criterion for such simulations. All the distances in Fig.~\ref{fig:compare} are sorted according to the image distances of the voxel intensity based simulation (intensity). The distances were zero-meaned and rescaled  to make the individual differences identifiable.  In Fig.~\ref{fig:compare}, we compared the longitudinal (long) and cross-sectional (cross) simulations as well as the combined weights of both (combine) respectively according to the P-B distances (Fig.~\ref{fig:compare}-(a)) and the P-rA distances (Fig.~\ref{fig:compare}-(b)). It is noticeable that the longitudinal information (long) outperformed the cross-sectional  information (cross) in most cases. Since neither method achieved the lowest simulation errors in all trials, the combined weights (combine) could be used to balance two perspectives. Taking the intensity-based method (intensity) as a reference, at least one of the proposed morphometry-based methods (long, cross, combine) achieved lower simulation errors in most cases regarding both criteria.

\section{Conclusion}
We present a framework which automatically simulates  future neurodegeneration from the longitudinal atrophic changes as well as the cross-sectional difference from a statistical average normal template. This framework expects the patient to have at least 2 historical MR images to increase the confidence of the simulated 3D MR volume. The brain tissue deformation was represented by the voxel-based morphometry (VBM). Our evaluation showed that the intra-subject longitudinal information enhances the simulation accuracy. The  results from at least one of the proposed morphometry-based methods outperformed the state-of-the-art intensity-based method in most evaluated cases. With a sufficient template collection, our proposed framework can be used for validating the prediction made by neuroimaging measurements extracted from MR data.

\section*{Acknowledgment}
This work was supported by ARC, AADRF, NA-MIC (NIH U54EB005149), and NAC (NIH P41RR013218).

\ifCLASSOPTIONcaptionsoff
  \newpage
\fi

\bibliographystyle{IEEEtran}

\end{document}